\documentclass{article}
\usepackage{spconf,amsmath,epsfig, multirow}
\usepackage{times}
\usepackage{epsfig}
\usepackage{graphicx}
\usepackage{pifont}
\usepackage{stfloats}
\usepackage{multirow}
\usepackage{tabularx}

\title{GeoBind: Binding  text, image, and audio through satellite images}

\name{ Aayush Dhakal\textsuperscript{a}, Subash Khanal\textsuperscript{a}, Srikumar Sastry\textsuperscript{a}, Adeel Ahmad\textsuperscript{a,b}, Nathan Jacobs\textsuperscript{a}}

\address{\textsuperscript{a} Washington University in St. Louis - St. Louis, MO, USA \\
\textsuperscript{b} Taylor Geospatial Institute - St. Louis, MO, USA}

\begin{document}
%
\maketitle
\let\thefootnote\relax\footnotetext{Personal use of this material is permitted. However, permission to reprint/republish this material for advertising or promotional purposes or for creating new collective works for resale or redistribution to servers or lists, or to reuse any copyrighted component of this work in other works, must be obtained from the IEEE.}
\begin{abstract}
In remote sensing, we are interested in modeling various modalities for some geographic location. Several works have focused on learning the relationship between a location and type of landscape, habitability, audio, textual descriptions, etc. Recently, a common way to approach these problems is to train a deep-learning model that uses satellite images to infer some unique characteristics of the location. In this work, we present a deep-learning model, GeoBind, that can infer about multiple modalities, specifically text, image, and audio, from satellite imagery of a location. To do this, we use satellite images as the binding element and contrastively align all other modalities to the satellite image data. Our training results in a joint embedding space with multiple types of data: satellite image, ground-level image, audio, and text.
Furthermore, our approach does not require a single complex dataset that contains all the modalities mentioned above. Rather it only requires multiple satellite-image paired data. While we only align three modalities in this paper, we present a general framework that can be used to create an embedding space with any number of modalities by using satellite images as the binding element. Our results show that, unlike traditional unimodal models, GeoBind is versatile and can reason about multiple modalities for a given satellite image input.  

\end{abstract}
\begin{keywords}
Deep Learning, Multimodal Learning, Contrastive Learning
\end{keywords}
\section{Introduction}
\label{sec:intro}
Inferring about various attributes associated with specific geographic locations is an important task in remote sensing. Previous research efforts have predominantly focused on establishing connections between a location and a single characteristic such as land use, livability metrics, audio features, ground-level scenery, and textual descriptions~\cite{yurui2020towards, zhu2022land,khanal2023learning, basu2021investigating,dhakal2023sat2cap}. This has led to the development of deep learning models that infer some unique characteristics of a location given a corresponding satellite image~\cite{greenwell2018goes, zang2021land, 9323706, sastry2024birdsat, klemmer2023satclip}.
This study aims to advance the current landscape by developing a unified embedding space that seamlessly links multiple modalities with geolocation. Our key contribution lies in creating a singular embedding space, which can be used to infer different properties of a location using satellite imagery. However, this can be challenging due to several factors. Primarily, training such a model through a traditional deep-learning approach requires high dimensional data that spans across all these modalities. For example: to create an embedding space that relates satellite images with text, audio, and ground-level images, you would need to create a quadruplet dataset with co-located text, audio, ground-level image, and satellite image. If every data point needs to contain information about every single modality, as the number of modalities grows, it becomes extremely infeasible to collect such data. 

\begin{figure}[t!]
    \centering
    \includegraphics[width=1\linewidth]{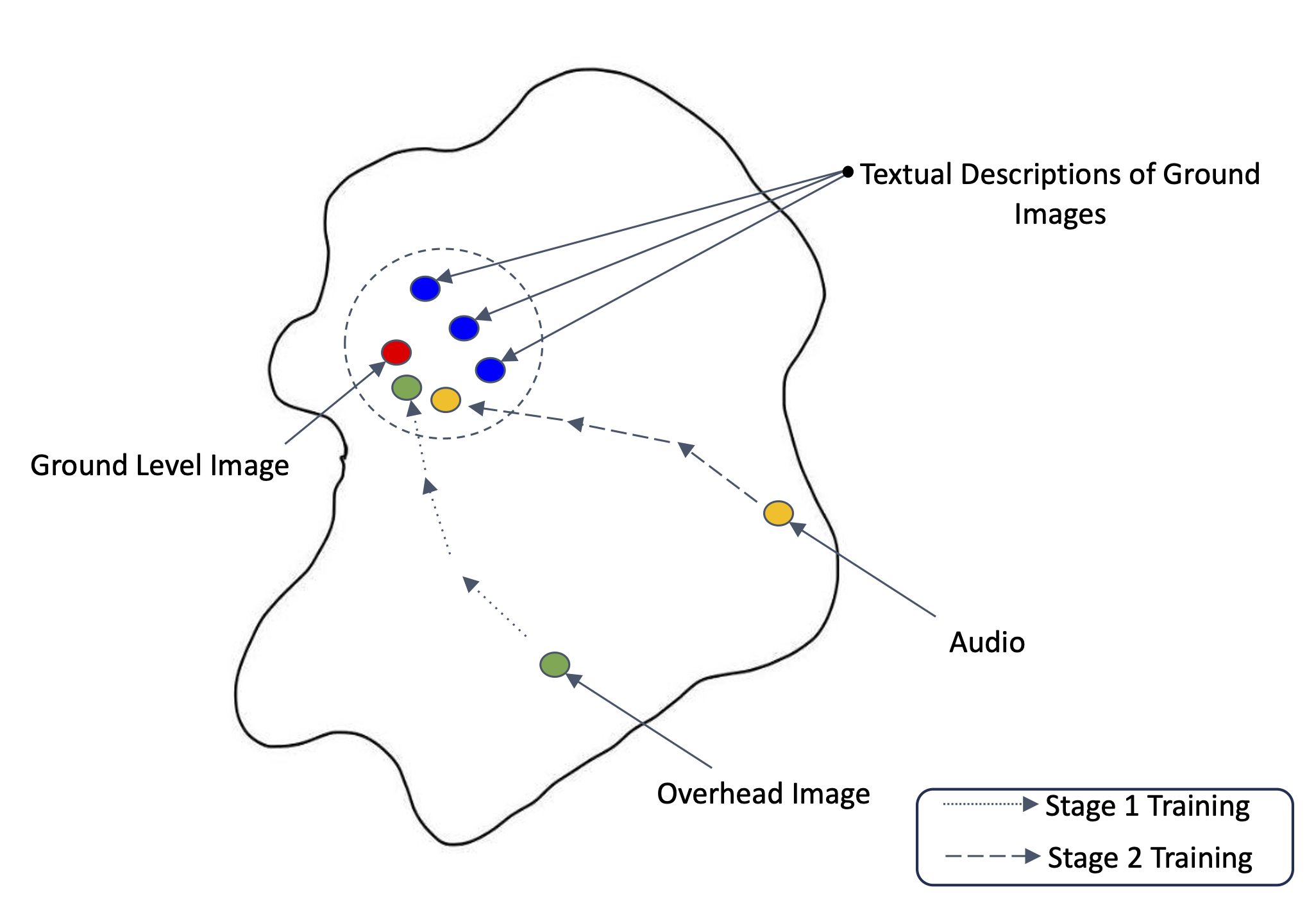}
    \caption{Using CLIP space, our framework creates a joint embedding space where semantically related satellite images, ground-level images, audio, and text data are pushed close together}.
    \label{fig:embedding_space}
\end{figure}

Recent work~\cite{girdhar2023imagebind} has addressed this issue by showing that it is possible to learn a joint embedding space for multiple modalities by using images to bind them all. ImageBind~\cite{girdhar2023imagebind} uses multiple image-paired datasets where each dataset consists of an image paired with some specific data type such as audio, video, etc. ImageBind then aligns each modality's embedding to the image embeddings, resulting in a shared representation space with all the modalities from different datasets. Inspired by this, we propose a framework that uses multiple satellite image-paired data to learn a joint embedding space where various modalities are bonded to a common representation space. Such embedding space would be beneficial to solve a wide variety of geospatial tasks. In this paper, we use two types of data: audio paired with satellite images, and ground-level images paired with overhead images. We use satellite imagery as a common medium to bind the different modalities. Our training occurs in two stages. In the first stage, we contrastively align the satellite images with ground-level images following Sat2Cap~\cite{dhakal2023sat2cap}. More specifically, the satellite image embeddings are aligned with the CLIP~\cite{radford2021learning} embeddings of the corresponding ground-level images. CLIP space already aligns semantically related text and images. Using the CLIP embeddings for alignment, we automatically align the satellite image embeddings with the textual descriptions of ground-level scenery as shown in Figure~\ref{fig:embedding_space}. In the second stage, we contrastively align the audio embeddings with satellite image embeddings using our model trained in stage 1. This results in a final joint embedding space, where semantically related satellite imagery, ground-level imagery, audio, and text are pushed together. Figure~\ref{fig:embedding_space} shows how the two stages of training result in a joint embedding space with four modalities. 

Our primary contribution is the introduction of a framework that allows us to create an embedding space with any ``n" number of modalities by using satellite imagery to bind them. Adding a new modality to an existing embedding space only requires adding a new stage to the process, which makes the framework easily scalable. In this paper, we show that with a minimal realizable model using 3 modalities. Our retrieval results demonstrate that our model, GeoBind, can perform geospatial tasks spanning different data types. Additionally, such a framework also results in the creation of emergent properties between the modalities which is extensively explored in ImageBind~\cite{girdhar2023imagebind}. For example, although there is no explicit training between satellite images and text, such a framework creates an emergent relationship between the two data types~\cite{dhakal2023sat2cap}. Overall, our contribution is the exploration of frameworks that move away from restrictive models that solve a single geospatial task and move towards versatile models capable of solving multiple tasks, without the requirement of highly complicated datasets. 

\section{Method}
\label{sec:method}
\subsection{Dataset}
We use two different satellite image-paired datasets in our work. First, we use the dataset from~\cite{dhakal2023sat2cap} which contains 6.1M pairs of overhead and ground-level images. The overhead images have a resolution of $0.6m/px$ and are acquired from Bing Maps. Each overhead image has a size of 800x800 px. Secondly, we use the SoundingEarth~\cite{wu2023large} data which contains 50k geo-tagged audio. Each of the audio samples is paired with a co-located overhead image with a resolution of $0.6m/px$. Each overhead image is 800x800 px and was acquired from Bing Maps. 

\subsection{Approach}
Our method has two steps of training. The first step aligns overhead images with ground-level images and consequently text data. The second step aligns audio data with satellite image embeddings resulting from the previous stage, which is shown in Figure~\ref{fig:embedding_space}.

In our first step, we align the overhead images with ground-level images in the CLIP space. For this step, we follow the procedure from the Sat2Cap~\cite{dhakal2023sat2cap} paper. For a batch of ground-level images $G_i$, we obtain their CLIP embeddings $C_i$. We have a Satellite Encoder, which takes a batch of satellite images $S_i$ and returns embeddings $O_i$. Now we contrastively align the satellite image embeddings with the corresponding CLIP embeddings using InfoNCE loss:
\begin{equation}
    L = \frac{1}{k}\sum_{i=0}^{k} -\log\frac{exp(o_i \cdot c_i / \tau)}{\sum_{j=0}^{k} exp(o_i \cdot c_j / \tau) }
\end{equation}
, where $\tau$ is the temperature parameter and k is the batch size.
Since the CLIP space aligns semantically related natural images and text, this process also aligns the satellite images with the textual descriptions of their ground-level scenery. At the end of this step, we get a Satellite Encoder, that takes as input a satellite image and projects it into CLIP space such that the resulting embedding is close to its co-located ground-level image and its textual descriptions in that space.

\begin{figure}[t!]
    \centering
    \includegraphics[width=1\linewidth]{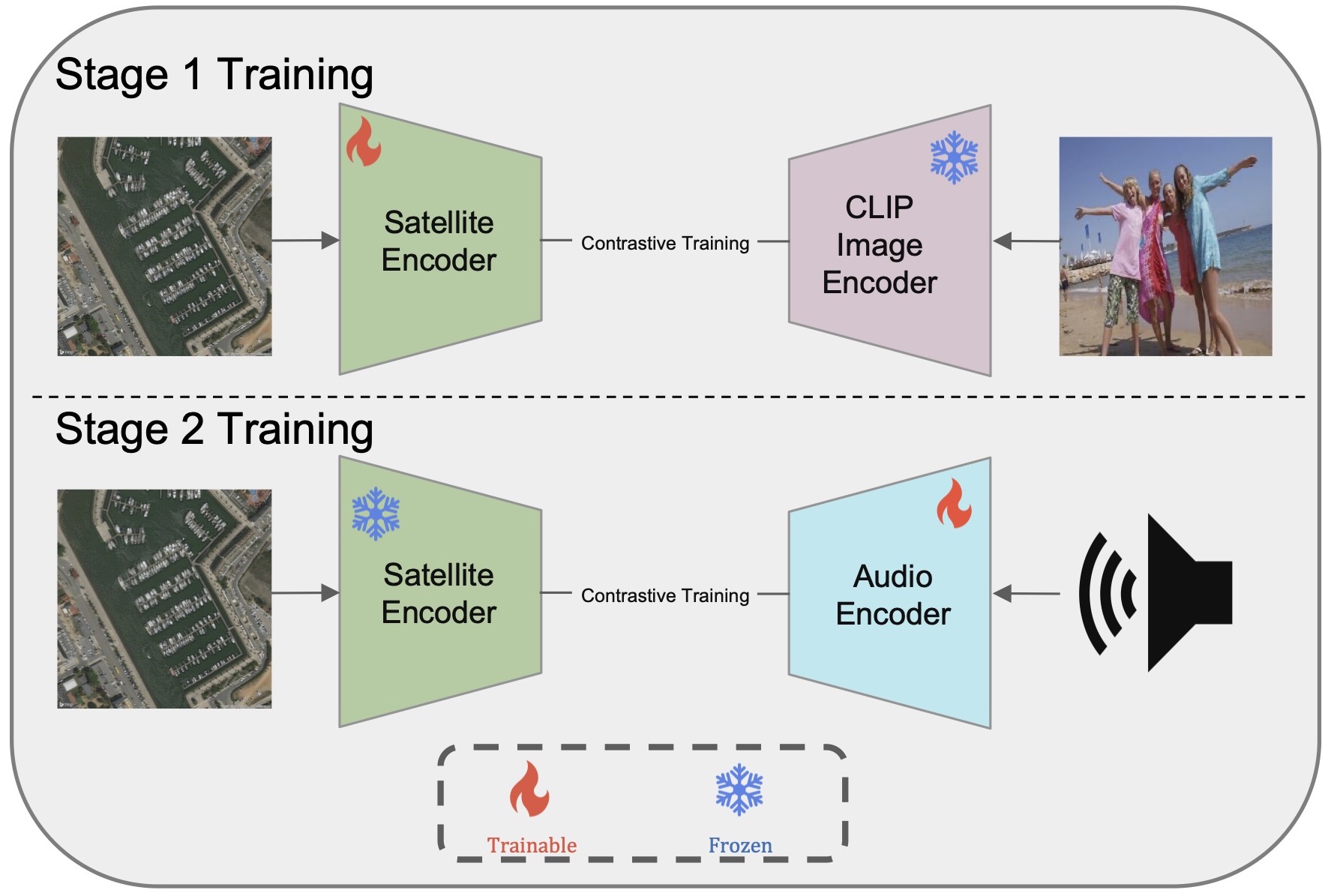}
    \caption{We employ a two-stage training framework. In the first stage, we contrastively update the Satellite Encoder while keeping the CLIP Image Encoder frozen. In the second stage, we contrastively update the Audio Encoder while keeping the Satellite Encoder frozen. It is important to note that more stages can be trivially added to this framework.}
    \label{fig:framework}
\end{figure}

In the second stage, our goal is to align audio embeddings with the satellite embeddings from stage 1. To achieve this, we utilize the SoudingEarth data, which has audio paired with satellite images. For a given batch of satellite images $S_i$, we compute their respective embeddings $O_i$ using the trained Satellite Encoder from stage 1. We now initialize an Audio Encoder, which takes as input a batch of audio recordings $H_i$ and returns a batch of audio embeddings $A_i$. We keep the Satellite Encoder frozen and contrastively train the Audio Encoder, such that, the audio embeddings move close to their corresponding satellite image embeddings. Our entire training framework is shown in Figure~\ref{fig:framework}.
\begin{equation}
    L_1 = \frac{1}{k}\sum_{i=0}^{k} -\log\frac{exp(o_i \cdot a_i / \tau)}{\sum_{j=0}^{k} exp(o_i \cdot a_j / \tau) }
\end{equation}

\begin{equation}
    L_2 = \frac{1}{k}\sum_{i=0}^{k} -\log\frac{exp(a_i \cdot o_i / \tau)}{\sum_{j=0}^{k} exp(a_i \cdot o_j / \tau) }
\end{equation}

\begin{equation}
    L = \frac{(L_1 + L_2)}{2}
\end{equation}

As the audio embeddings move closer to their corresponding satellite image embeddings, they also naturally align with semantically related ground-level images and textual captions as shown in Figure\ref{fig:embedding_space}. This creates a joint embedding space where different modalities can interact with each other. More specifically, we can use the Satellite Encoder, Audio Encoder, CLIP Image Encoder, and CLIP Text Encoder to project satellite image, audio, ground-level image, and text respectively, into a common space. The existence of such joint space allows us to perform tasks that require any permutation or combination of the given modalities eliminating the need for a single task-specific model. Furthermore, it is trivial to add additional stages to align other modalities with satellite imagery and project them into the same embedding space. Hence, the utility of this framework extends beyond the application we demonstrate in our work.

\section{Experiments and Results}
\label{sec:exp}
\subsection{Implementation Details}
We use a pretrained CLIP ViT-32B to generate the CLIP embeddings. We also use a ViT-32B as the Satellite Encoder and initialize it with the CLIP model parameters. We use the CLAP audio encoder from Hugging Face to initialize our audio encoder. We use RandAugment~\cite{cubuk2020randaugment} with 3 operations to augment our satellite images during training. We use AdamW~\cite{loshchilov2017decoupled} optimizer with a learning rate of $5e-05$ with $\beta_1=0.99$ and $\beta_2=0.98$. We also use CosineAnnealing with Warm Restarts~\cite{loshchilov2016sgdr} as the learning rate scheduler. The temperature parameter was set to be learnable during training and was initialized to $0.07$.   

\subsection{Crossmodal Retrieval}
To show that our embedding space aligns semantically related data together, we perform retrieval experiments. More specifically, we project two different data types into our joint embedding space and compute the cosine similarity between them. All the experiments are carried out using a held-out test set with 10,000 samples. Firstly, we show our embedding space captures the relation between co-located satellite images and ground-level images. We compute the embedding for satellite images using our Satellite Encoder and use the CLIP Image Encoder to compute the embeddings for the ground-level images. We then calculate the cosine similarity between all possible pairs and compute top-k metrics. Since stage 1 training is identical to Sat2Cap~\cite{dhakal2023sat2cap}, we see identical retrieval performance when not providing any metadata to the model in Table~\ref{table:image_retrieval}. As baseline, we use the CLIP embeddings of the satellite-image as in~\cite{dhakal2023sat2cap}. The results in Table~\ref{table:image_retrieval} indicates, about 56\% of the time, the ground-truth image lies within the top-10 when ranked according to their cosine similarity. This tells us that semantically related satellite and ground-level images are aligned in our embedding space. Since we are optimizing over the CLIP space, naturally our overhead image embeddings also move close to the textual descriptions of the regions, as shown in Figure~\ref{fig:embedding_space}. Sat2Cap~\cite{dhakal2023sat2cap} shows qualitative and quantitative results demonstrating that this type of training eventually leads to aligning the satellite images with textual descriptions of the ground-level scenery.

\newcommand{\cmark}{\ding{51}}%
\newcommand{\xmark}{\ding{55}}%

\begin{table}[htpb]
\centering
\begin{tabular}{l|c|c}
\hline
\multicolumn{1}{c|}{Model} & \multicolumn{1}{l|}{Recall@10} & \multicolumn{1}{l}{Median Rank} \\ \hline
CLIP           & 1.3           & 2857          \\ \hline
Sat2Cap        & \textbf{56.4} & \textbf{13.5} \\ \hline
GeoBind (ours) & \textbf{56.4} & \textbf{13.5}
\end{tabular}
\caption{\textbf{Top-k Sat2Image Retrieval Scores:} We compare the satellite-to-ground image retrieval performance of our model. Since the stage1 training is identical to Sat2Cap, we get identical performance without the use of location and time metadata.}
    \label{table:image_retrieval}
\end{table}

Secondly, we evaluate the satellite image to audio retrieval performance of our model. To do this, we compute the satellite embeddings using our Satellite Encoder and audio embeddings using our Audio Encoder. Similar to the previous experiment, we compute the top-k metrics, which are shown in Table~\ref{table:sound_retrieval}. Firstly, we notice that the scores are much worse than satellite-image to ground-image retrieval. However, this happens because satellite-to-audio retrieval is inherently a much harder task due to the ill-posed nature of the problem. This issue has been discussed in further detail in the SoundingEarth~\cite{heidler2023self} paper. Table~\ref{table:sound_retrieval} shows that while we do not achieve the highest recall, our scores are comparable to the existing models for this task. The results indicate that our framework trades-off accuracy for versatility. While the SoundingEarth satellite-image encoders are limited to only reasoning about audio data, our satellite-image encoder can reason about multiple modalities like ground-level images, audio, and text. 

The retrieval results confirm that GeoBind creates a joint embedding space by aligning semantically related satellite image, audio, and ground-level image (and thereby text due to the image-text alignment of CLIP space). Hence, the GeoBind framework gives us a single satellite-image encoder that can reason about different types of data, effectively eliminating the need for distinct deep-learning models for each modality. Furthermore, we achieve this versatility while maintaining comparable performance to the modality-specific models. Pushing towards such models would make it much easier to work across modalities as well as work on tasks that require some combination of multiple data types.
\begin{table}[htbp]
\footnotesize
\begin{tabular}{l|l|l|l}
\hline
           Model    & Loss                & Recall@100 & Median Rank \\ \hline
               & Native TL & 18.59  & 749                        \\
SoundingEarth~\cite{wu2023large}  & Contrastive    & \textbf{29.12} & \textbf{565}                 \\
               & Batch TL  & 19.01 & 744                         \\ \hline
GeoBind (ours) &     & \underline{24.63} & \underline{613}   

\end{tabular}
    \caption{\textbf{Top-k Sat2Audio Retrieval Scores:} We compare the satellite image-to-sound retrieval performance of our model with previous models. Results show that the performance of GeoBind is comparable to other similar models although it is trained with multiple modalities.}
    \label{table:sound_retrieval}
\end{table}

\section{Discussion and Conclusion}
In our study, we presented a framework that allows satellite images to interact with multiple types of data. By binding multiple modalities to satellite imagery, we created a joint embedding space that combines semantically related text, ground-level images, audio, and satellite images. The resulting joint embedding space can be further fine-tuned to solve more specific problems as per the requirements of the user.

Our main goal in this paper is to encourage the development of general-purpose deep-learning models that can reason about multiple characteristics of a given satellite image data. While we used a two-stage training process, it is possible to add more stages to our framework and train with a larger variety of data types. By aligning new datatypes through satellite images, we can create a joint embedding space with any number of modalities, all bound through geolocation. Our work motivates the development of a small number of versatile and efficient models, as opposed to a large number of highly specific unimodal models. In future work, we would like to explore the addition of more modalities using this framework as well as conduct studies on emergent properties.
\bibliographystyle{IEEEbib}
\bibliography{strings,refs}

\end{document}